%% file: main.tex
\documentclass[10pt,twocolumn,letterpaper]{article}

\usepackage[pagenumbers]{cvpr} 

\input{preamble}

\definecolor{cvprblue}{rgb}{0.21,0.49,0.74}
\usepackage[pagebackref,breaklinks,colorlinks,allcolors=cvprblue]{hyperref}

\def\paperID{24196} 
\def\confName{CVPR}
\def\confYear{2027}

\title{Angular Conditional Distillation: Overcoming Classifier Geometry Mismatch}

\author{
  Xuefei Zhang\textsuperscript{1,*} \quad 
  Ruoqi Wang\textsuperscript{1} \quad 
  Lei Cui\textsuperscript{1,\dag} \\
  \textsuperscript{1}Institute of AI for Industries (IAII), Chinese Academy of Sciences \\
  \texttt{\{xfzhang, rqwang, lcui\}@iaii.ac.cn} \\
  \textsuperscript{*}First author \quad \textsuperscript{\dag}Corresponding author
}

\begin{document}
\maketitle
\input{sec/0_abstract}    
\input{sec/1_intro}
\input{sec/2_rw}

\input{sec/3_analy}
\input{sec/4_mth}

\input{sec/5_exp}
\input{sec/6_conclusion}
{
    \small
    \bibliographystyle{ieeenat_fullname}
    \bibliography{20261110}
}


\end{document}

%% file: preamble.tex
\usepackage{amsmath}
\usepackage{algorithm}
\usepackage{algpseudocode}

\usepackage{booktabs}
\usepackage{multirow}
\usepackage{array}
\usepackage{graphicx}

%% file: sec/0_abstract.tex
\begin{abstract}

Classifier outputs provide a fundamental and universally available knowledge interface for knowledge distillation (KD). However, standard logit objectives entangle transferable semantic relations with architecture-dependent parameterization artifacts, causing a classifier geometry mismatch that makes direct knowledge transfer unreliable.
To address this, we propose Angular Conditional Distillation (ACD). First, the Angular Coordinate Interface (ACI) reconstructs the student classifier using angular coordinates, suppressing magnitude and bias effects to provide a geometry-compatible space. Then, Scale-Invariant Conditional Distillation (SICD) standardizes outputs and exclusively transfers conditional non-target relations, filtering out untransferable absolute confidence. Together, ACI defines where to transfer, while SICD defines what to transfer.
Extensive experiments show that ACD achieves state-of-the-art performance over existing KD methods. Moreover, as a universal drop-in replacement, it consistently improves existing distillation pipelines while requiring no feature adapters or additional inference-time cost. Code is available at \url{https://github.com/laniusnan/ACD}.

\end{abstract}

%% file: sec/1_intro.tex
\section{Introduction}
\label{sec:intro}

The continued scaling of neural networks has yielded remarkable advances across several tasks \cite{simeoniDINOv32025, zhaoDETRsBeatYOLOs2024, raviSAM2Segment2024}, yet their high computational costs hinder deployment on edge devices \cite{mansourianComprehensiveSurveyKnowledge2025}. Knowledge distillation (KD) addresses this by transferring knowledge from a heavy teacher to a lightweight student. Over the past decade, KD has evolved from matching classifier outputs to leveraging intermediate features \cite{romeroFitNetsHintsThin2015}, sample relations \cite{tianContrastiveRepresentationDistillation2022}, and structural knowledge \cite{pengCorrelationCongruenceKnowledge}.
Nevertheless, classifier outputs remain the most universally available supervision interface. Unlike feature- or relation-based methods that demand architectural\cite{chenDistillingKnowledgeKnowledge} alignment or extra modules \cite{tianContrastiveRepresentationDistillation2022}, logits offer a lightweight, architecture-agnostic endpoint. They capture rich inter-class correlations and are highly flexible, serving either as the primary objective in pure logit KD \cite{zhaoDecoupledKnowledgeDistillation2022} or as a plugin auxiliary loss across diverse paradigms \cite{chenDistillingKnowledgeKnowledge}.

\begin{figure}[t]
  \centering
  \begin{minipage}[b]{0.60\linewidth}
    \centering
    \includegraphics[width=\linewidth]{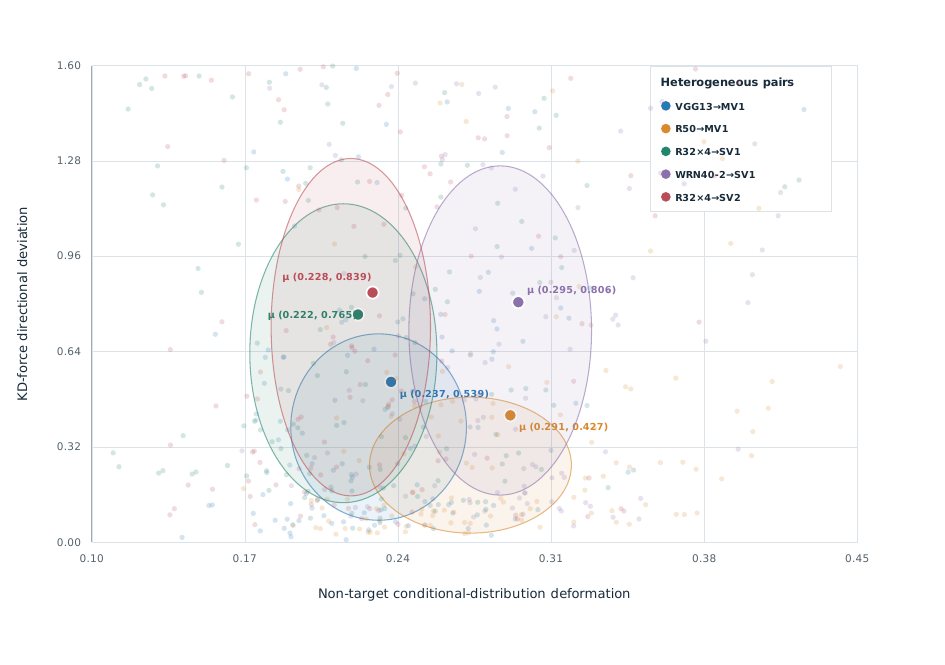}
    \centerline{\small (a) Empirical evidence}
  \end{minipage}
  \hfill 
  \begin{minipage}[b]{0.39\linewidth}
    \centering
    \includegraphics[width=\linewidth]{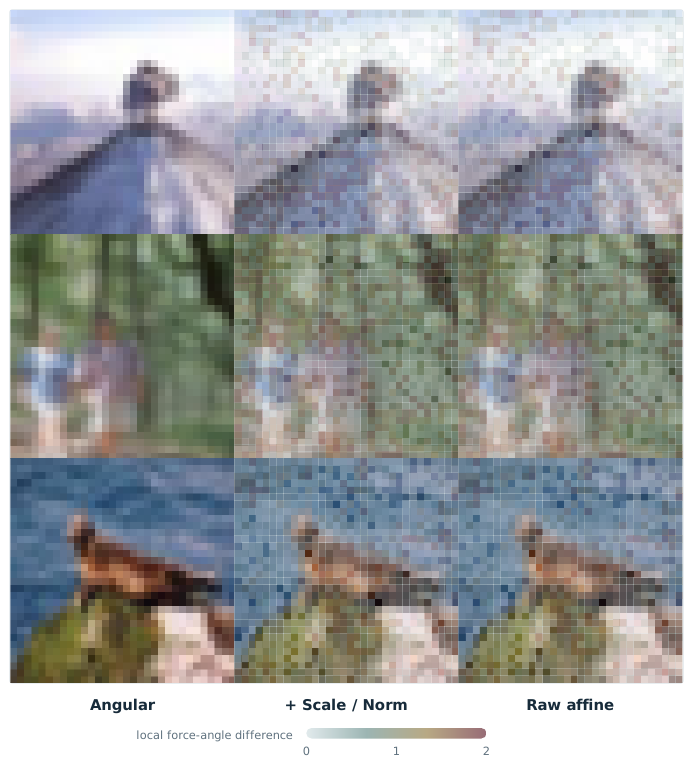}
    \centerline{\small (b) Directional disagreement}
  \end{minipage}
  
  \caption{\textbf{Classifier Geometry Mismatch.} 
  \textbf{(a)} Empirical evidence: fixing teacher/student features but restoring parameterization causes substantial deformation in non-target distributions (x-axis) and gradient deviations (y-axis). 
  \textbf{(b)} Local gradient directional disagreement under different readouts (VGG13\(\rightarrow\)MobileNetV1). Darker overlays indicate higher disagreement \(1-\cos(g_p,g_p^{\mathrm{ang}})\).}
  \label{fig:geometry_mismatch}
\end{figure}

However, conventional logit matching implicitly assumes that teacher and student outputs reside in compatible coordinates\cite{zhaoDecoupledKnowledgeDistillation2022}. We argue this assumption is fundamentally flawed.
For a linear classifier, the logit of class $k$ can be written as
$z_k=\mathbf{w}_k^{\top}\mathbf{h}+b_k
=\|\mathbf{w}_k\|\|\mathbf{h}\|\cos\theta_k+b_k$,
where $\mathbf{h}$ denotes the feature representation and $\theta_k$ is its angle with the class weight $\mathbf{w}_k$.
This decomposition reveals that a raw logit entangles transferable directional relations (\(\theta_k\)) with parameterization artifacts (feature magnitudes, weight norms, and biases). Consequently, classifiers with identical  directions can yield entirely different supervisory signals. We refer to this discrepancy as \emph{classifier geometry mismatch}.

To empirically verify this issue, we conduct a controlled probe (Fig.~\ref{fig:geometry_mismatch}).
Across five heterogeneous pairs, merely restoring affine parameterization induces substantial distribution deformation and severe gradient shifts. 
Crucially, this mismatch is not merely a scalar confidence error. Even with oracle-calibration in Sec.~\ref{sec:analysis}, that applying the optimal positive scalar to student logits to eliminate confidence gaps, substantial residual mismatch remains. This confirms that traditional KD forces students to reproduce parameterization artifacts. This raises a central question: How can we transfer informative class relationships while minimizing dependence on parameterization?

Recent logit-based approaches have improved output supervision via decoupling\cite{zhaoDecoupledKnowledgeDistillation2022} or normalization\cite{chiNormKDNormalizedLogits}. However, they primarily modify how information is weighted after the classifier has defined its coordinates, failing to address whether the coordinates are compatible and which relations are transferable. 

To bridge this gap, we propose \emph{Angular Conditional Distillation} (ACD). We introduce the \emph{Angular Coordinate Interface} (ACI), which reconstructs the student classifier using normalized angular coordinates (\(\tilde{z}_k = s\cos\theta_k\)). By removing magnitudes and biases, ACI provides a geometry-compatible space. Furthermore, we initialize the classifier weights with an Equiangular Tight Frame (ETF) to provide an optimal angular prior. During training, this allows data-sensitive classes to refine their specific directions, while data-insensitive classes maintain their geometrically optimal, maximally separated angles, ensuring a robust and structured representation space. 

However, establishing an angular space is insufficient. If standard KL is applied to ACI, it still forces the student to replicate the teacher's absolute confidence, which is highly sensitive to the capacity gap. To address this, we propose \emph{Scale-Invariant Conditional Distillation} (SICD). Specifically, SICD first applies dual-side standardization to both teacher and student outputs to eliminate residual scale and shift effects. Then, using a finite-temperature KL formulation, SICD mathematically decomposes the divergence into a target-confidence term and a conditional non-target relation term. By dropping the target-confidence term and exclusively optimizing the latter, SICD transfers only the true dark knowledge (relative class relations) without imposing untransferable absolute confidence. Furthermore, a gradient norm matching term is employed to stabilize the optimization scale. 

Together, ACI defines the compatible geometric space (\emph{where} to transfer), while SICD isolates and transfers the pure relational dark knowledge (\emph{what} to transfer).
In summary, our contributions are threefold:
\begin{itemize}
\item \textbf{Problem Identification.} We identify \emph{classifier geometry mismatch} as a fundamental obstacle in logit distillation, revealing how transferable directional class relations are deeply entangled with untransferable parameterization artifacts across diverse architectures.
\item \textbf{Methodology.} We propose ACD, a geometry-aware framework consisting of two synergistic modules. \emph{ACI} establishes a well-conditioned angular coordinate space augmented by an ETF initialization, while \emph{SICD} mathematically decomposes the KL objective to transfer exclusively the scale-invariant conditional non-target relations.
\item \textbf{Empirical Validation.} We demonstrate that ACD achieves state-of-the-art results across various settings. Notably, it serves as a universal replacement for conventional logit matching, consistently boosting existing KD pipelines while maintaining the extreme efficiency of logit-only methods.
\end{itemize}


%% file: sec/2_rw.tex
\section{Related Work}
\label{sec:related_work}
\begin{algorithm}[t]
\caption{Per-Sample Oracle Calibration Analysis}
\label{alg:oracle_calibration}
\small
\begin{algorithmic}[1]
\Require Frozen teacher $f_t$ and student $f_s$
\Require Images $\{x_i\}_{i=1}^{N}$; temperature $T>0$
\Ensure Per-image scales and KL divergences before/after calibration

\For{$i=1,\ldots,N$}
    \State $z_i^t \gets f_t(x_i)$,
           $z_i^s \gets f_s(x_i)$
    \State $p_i^t \gets \operatorname{softmax}(z_i^t/T)$
    \State Define $p_i^s(\alpha)
           = \operatorname{softmax}(\alpha z_i^s/T)$
    \State Define $\ell_i(\alpha)
           = T^2 D_{\mathrm{KL}}
             \bigl(p_i^t \,\|\, p_i^s(\alpha)\bigr)$
    \State $\alpha_i^\star
           \gets \underset{\alpha\geq 0}{\arg\min}\;
           \ell_i(\alpha)$
    \State $e_i^{\mathrm{before}} \gets \ell_i(1)$
    \State $e_i^{\mathrm{after}} \gets \ell_i(\alpha_i^\star)$
\EndFor

\State \Return
$\{\alpha_i^\star,
   e_i^{\mathrm{before}},
   e_i^{\mathrm{after}}\}_{i=1}^{N}$
\end{algorithmic}
\end{algorithm}

\begin{figure}[t]
    \centering
    \begin{subfigure}[t]{0.48\linewidth}
        \centering
        \includegraphics[width=\linewidth]
        {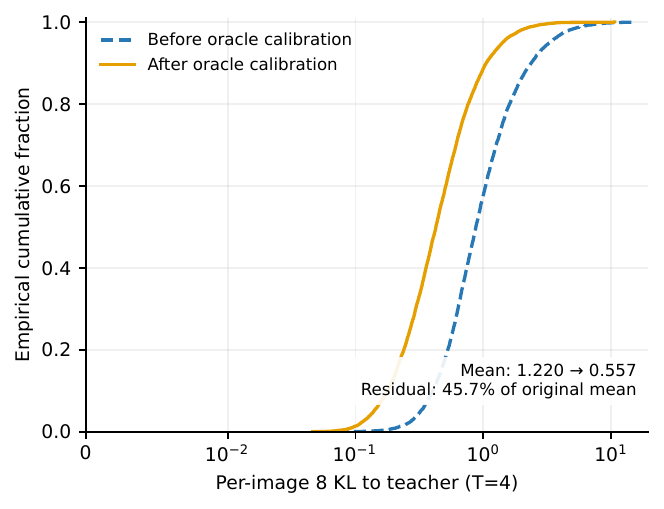}
        \caption{Empirical distributions}
        \label{fig:output_mismatch_a}
    \end{subfigure}
    \hfill
    \begin{subfigure}[t]{0.48\linewidth}
        \centering
        \includegraphics[width=\linewidth]
        {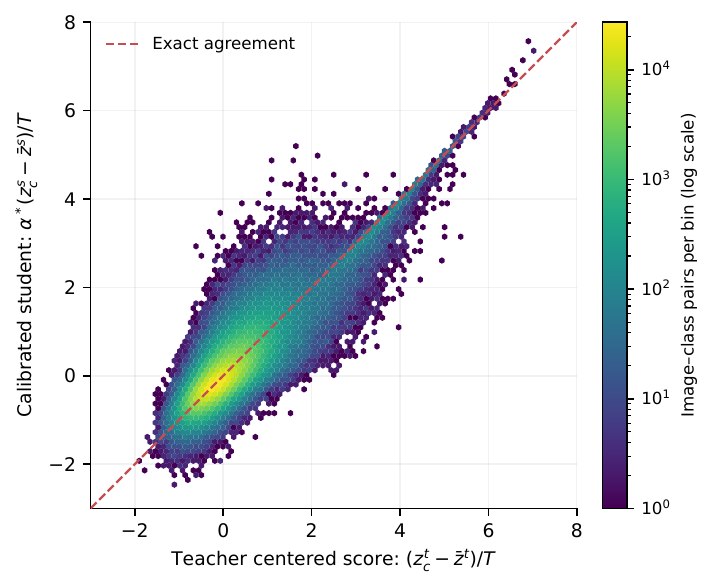}
        \caption{Density}
        \label{fig:output_mismatch_b}
    \end{subfigure}

    \caption{\textbf{Output mismatch beyond scalar confidence.}
    (a) Empirical cumulative distributions of the weighted KL
    divergence before and after per-image
    oracle calibration, with a symmetric-log horizontal axis.
    (b) Density of image-class pairs, comparing centered teacher
    scores with calibrated, centered student scores, both divided
    by $T$.}
    \label{fig:output_mismatch}
\end{figure}

\begin{figure*}[t]
\centering
\includegraphics[width=\textwidth]{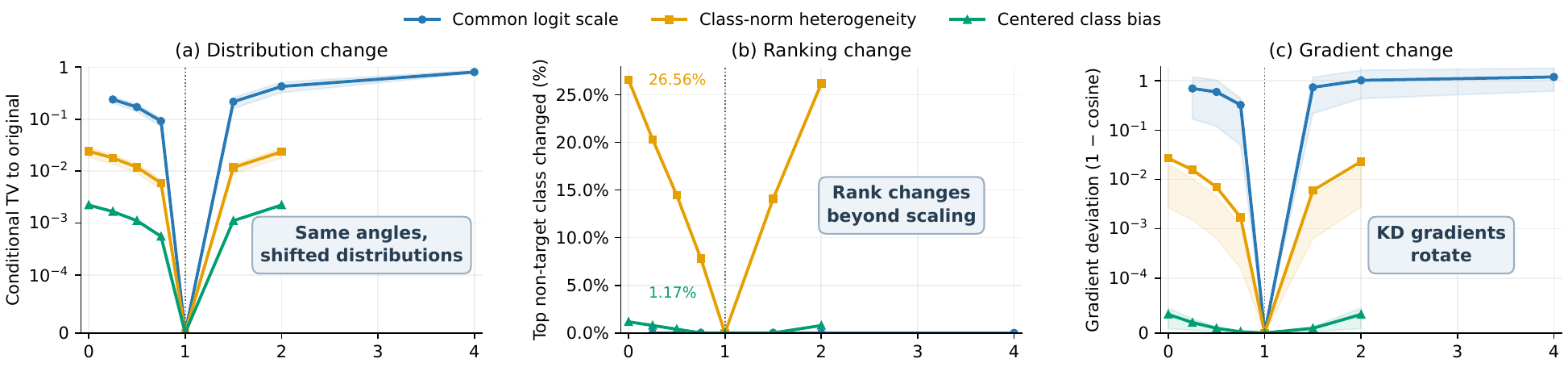}
\caption{\textbf{Sensitivity to score parameterization.}
Panels show (a) non-target conditional total variation at $T=4$,
(b) the fraction of images whose highest-scoring non-target
class changes, and (c) KD feature-gradient direction deviation
($1-\cos$), all relative to the original model.}
\label{fig:parameterization_factors}
\end{figure*}

\begin{figure}[t]
  \centering
  \includegraphics[width=\linewidth]{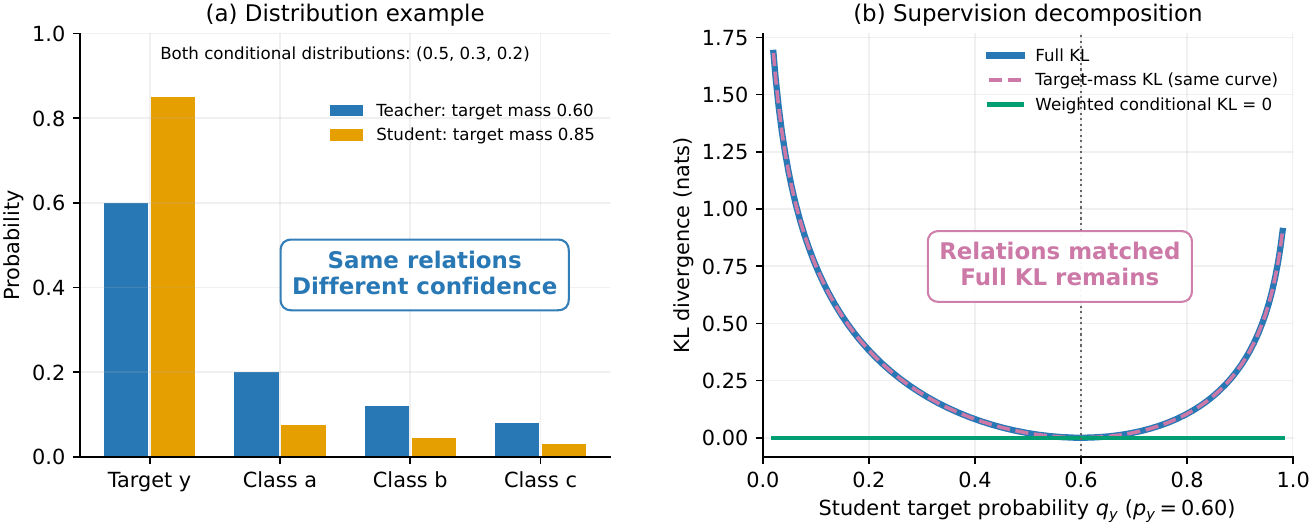}
  \caption{\textbf{Relation distribution.}
  (a) Teacher and student have different target probabilities, but identical non-target
  conditional distributions.
  (b) Varying $q_y$ while fixing these conditional
  distributions leaves the weighted conditional KL at zero.
  Full KL equals the target-mass term and remains positive
  unless $q_y=p_y$.}
  \label{fig:supervision_problem}
\end{figure}

Beyond output matching, knowledge distillation has been explored
through intermediate features~
\cite{romeroFitNetsHintsThin2015, yuanRevisitingKnowledgeDistillation,heoComprehensiveOverhaulFeature},
representation relations~
\cite{pengCorrelationCongruenceKnowledge,tianContrastiveRepresentationDistillation2022,tungSimilarityPreservingKnowledgeDistillation2019,
liuKnowledgeDistillationInstance2019}, and structured or heterogeneous
representation-transfer mechanisms~
\cite{joCodebookGuidedCrossModalKnowledge2026a,yuCoCaRSCorrelationCalibrationBased2026,etcibasiYouDontNeed2026}.

Our work focuses on class-level output distillation.
Classical KD matches temperature-softened teacher and student
distributions~\cite{hintonDistillingKnowledgeNeural2015}.
Subsequent studies examine its connection to label smoothing
regularization~\cite{yuanRevisitingKnowledgeDistillation}, reduce magnitude
dependence through logit standardization~\cite{sunLogitStandardizationKnowledge},
or introduce alternative discrepancy measures~
\cite{lvWassersteinDistanceRivals2024,wangABKDPursuingProper2025}.
DIST transfers correlations among predictions~
\cite{beyerKnowledgeDistillationGood}, while DKD separates target-class
and non-target-class supervision~\cite{zhaoDecoupledKnowledgeDistillation2022},
and NKD investigates non-target normalization~
\cite{chiNormKDNormalizedLogits}.
Recent studies in large language and multimodal models
further explore confidence-aware, uncertainty-aware,
and token-selective distillation~
\cite{senguptaDistillingWhatMatters2026,sunUncertaintyawareKnowledgeDistillation2026,kimExplainYourOwn2026}.

In contrast, ACD establishes an angular student interface to
separate directional class relations from magnitude and bias
effects, and transfers scale-invariant non-target conditional
relations without a separate target-confidence matching term.

%% file: sec/3_analy.tex
\section{Empirical Analysis}
\label{sec:analysis}

\subsection{Is the mismatch merely a confidence-scale difference?}
Figure~\ref{fig:geometry_mismatch} illustrates readout-dependent
changes in non-target distributions and local KD gradient
directions. To examine whether teacher--student output mismatch
can be explained by overall confidence scale alone, we perform
per-sample oracle calibration:
\begin{equation}
    \alpha_i^\star
    = \underset{\alpha \geq 0}{\arg\min}\;
      T^2 D_{\mathrm{KL}}
      \bigl(p_i^t \,\|\, p_i^s(\alpha)\bigr).
    \label{eq:oracle_calibration}
\end{equation}
where the teacher distribution $p_i^t$ and the calibrated
student distribution $p_i^s(\alpha)$ are defined as:
\[
\begin{aligned}
    p_i^t
    &= \operatorname{softmax}(z_i^t/T), \\
    p_i^s(\alpha)
    &= \operatorname{softmax}(\alpha z_i^s/T).
\end{aligned}
\]

Both models remain frozen, and each image receives its own
optimal scalar (Alg.~\ref{alg:oracle_calibration}), making
this diagnostic more flexible than a single global temperature.
Across all 10,000 CIFAR-100 test images, calibration reduces
the mean $T^2$-scaled KL divergence from 1.2198 to 0.5575,
leaving 45.7\% of the original discrepancy
(Fig.~\ref{fig:output_mismatch}(a)).
The calibrated, centered class scores also remain dispersed
around the identity line
(Fig.~\ref{fig:output_mismatch}(b)).
Thus, for this checkpoint, even image-specific scalar
calibration cannot fully account for the teacher--student
output mismatch, motivating a closer examination of
class-dependent score parameterization.

\subsection{Why an angular interface?}
A linear classifier entangles angular relations with magnitude-
and bias-dependent modulation:
\begin{equation}
    z_k
    = \|\mathbf{h}\|_2 \|\mathbf{w}_k\|_2
      \cos\theta_k + b_k,
    \label{eq:logit_decomposition}
\end{equation}
where $\theta_k$ is the angle between the student feature
$\mathbf{h}$ and the class-weight vector $\mathbf{w}_k$.
To isolate the effects of non-angular factors, we fix student
features, class-weight directions, and teacher outputs, and
separately vary the common logit scale, class-weight norm
heterogeneity, and centered class biases.
Despite unchanged angular relations, these interventions alter
non-target conditional distributions
(Fig.~\ref{fig:supervision_problem}a).
Class-dependent factors can additionally change the highest-scoring
non-target class, an effect that positive scalar calibration
cannot reverse (Fig.~\ref{fig:supervision_problem}b).
These changes also propagate to the direction of the local
KD feature gradient (Fig.~\ref{fig:supervision_problem}c).
Together, these observations motivate ACI: a normalized,
bias-free student readout that separates angular relations
from feature magnitudes, class-weight norms, and additive biases,
while retaining a shared output scale.

\begin{figure*}[t]
  \centering
  \includegraphics[width=\textwidth]{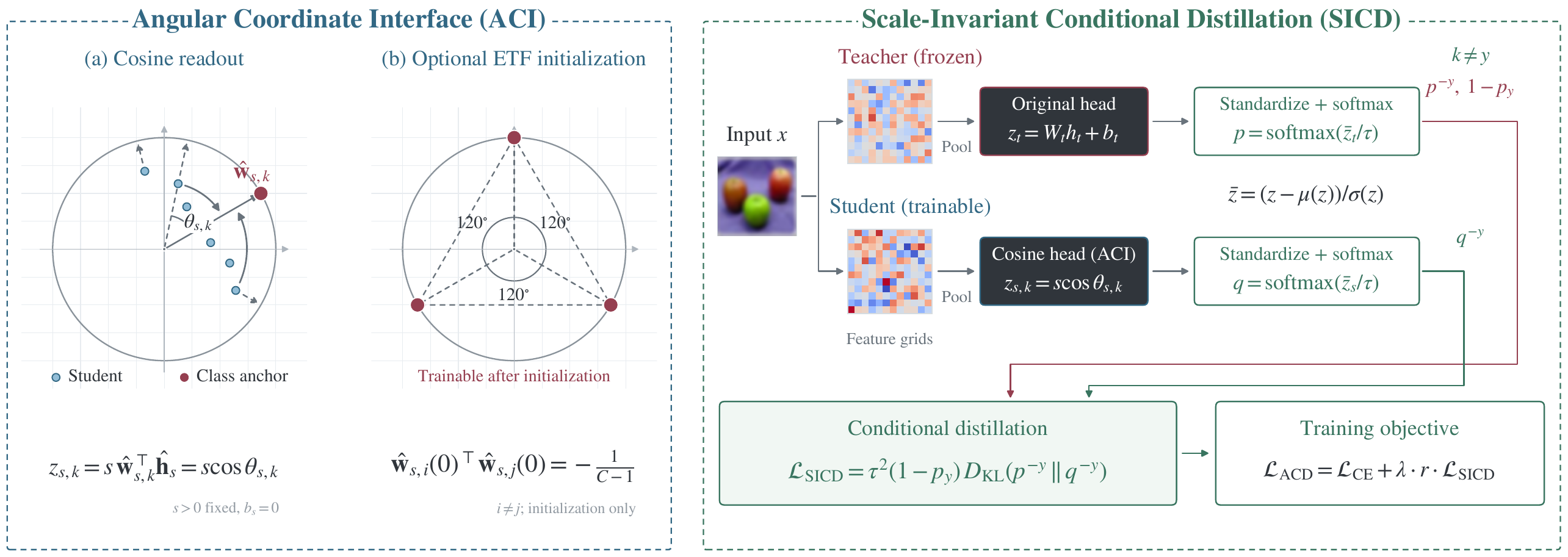}
  \caption{\textbf{Overview of ACI and SICD.}
  ACI parameterizes student logits using 
  cosine similarities with a shared scale, removing explicit
  feature magnitude, class-specific weight norms, and biases.
  The teacher retains its original classifier.
  SICD standardizes teacher and student logits and matches
  their non-target conditional distributions, weighted by
  the teacher's non-target probability mass.
  A detached gradient-strength matching factor calibrates
  the conditional objective against standardized full KL.
  The student is trained jointly with ground-truth
  cross-entropy supervision.}
  \label{fig:method_overview}
\end{figure*}

\subsection{What should be distilled within the angular interface?}

An angular interface specifies how class scores are
parameterized, but not which aspects of the teacher
distribution should be matched.
Let $p$ and $q$ denote the teacher and student distributions.
Their KL divergence decomposes as
\begin{equation}
\begin{aligned}
    D_{\mathrm{KL}}(p \| q)
    &= \underbrace{
        D_{\mathrm{KL}}^{\mathrm{bin}}(p_y \| q_y)
       }_{\text{target-mass matching}} \\
    &\quad + \underbrace{
        (1-p_y)D_{\mathrm{KL}}(p^{-y} \| q^{-y})
       }_{\text{non-target relation matching}},
\end{aligned}
\label{eq:conditional_kl_decomposition}
\end{equation}
where $D_{\mathrm{KL}}^{\mathrm{bin}}$ denotes binary KL,
and $p^{-y}$ and $q^{-y}$ are the conditional distributions
over non-target classes.

Figure~\ref{fig:supervision_problem} distinguishes relation
transfer from confidence matching.
Figure~\ref{fig:supervision_problem}~(a) shows that teacher and student can share identical
non-target conditional distributions, $p^{-y}=q^{-y}$,
while assigning different target probabilities.
Figure~\ref{fig:supervision_problem}~(b) varies only $q_y$: the weighted conditional term
remains zero, yet full KL is positive whenever $q_y\ne p_y$.
Thus, matching non-target relations does not remove
full KL's additional target-confidence supervision.
This controlled probability-space example motivates SICD
to retain the weighted conditional term while omitting
explicit target-mass matching, with cross-entropy providing
ground-truth supervision.



%% file: sec/4_mth.tex
\section{Method}
\label{sec:method}

In this section, we present \emph{Angular Conditional Distillation} (ACD), a geometry-aware framework designed to bridge the classifier geometry mismatch in output-based KD. ACD consists of two core components: (1) an \emph{Angular Coordinate Interface} (ACI) that reconstructs the student classifier into a pure angular space, eliminating parameterization artifacts like feature magnitudes and biases; and (2) a \emph{Scale-Invariant Conditional Distillation} (SICD) objective that transfers only the scale-invariant non-target relations. We first revisit the standard KD formulation, then detail ACI and SICD respectively.

\subsection{Preliminaries}
\label{subsec:preliminaries}
Knowledge Distillation (KD) involves using a fixed teacher model \(f^T\) to improve a parameterized student model \(f^S\). Given an input \(x\), the teacher and student produce logits \(\mathbf{z}^T\) and \(\mathbf{z}^S\), which are converted into probability distributions \(p = \sigma(\mathbf{z}^T/\tau)\) and \(q = \sigma(\mathbf{z}^S/\tau)\) via a softmax function \(\sigma\) and a temperature \(\tau\). The standard KD objective minimizes the divergence between these distributions:
\begin{equation}
    \mathcal{L}_{\text{KD}} = \tau^2 D_{\text{KL}}(p \parallel q).
\end{equation}
The final training loss is a combination of the cross-entropy loss \(\mathcal{L}_{\text{CE}}\) and the distillation loss:
\begin{equation}
    \mathcal{L}_{\text{total}} = \mathcal{L}_{\text{CE}} + \lambda \mathcal{L}_{\text{KD}},
\end{equation}
where \(\lambda\) is a balancing hyper-parameter.

As discussed, a raw linear logit \(z_k = \|\mathbf{w}_k\|\|\mathbf{h}\|\cos\theta_k + b_k\) deeply entangles transferable directions with untransferable magnitudes and biases. 

\begin{table*}[t]
\centering
\caption{Comparison of Top-1 accuracy on CIFAR-100. We report the individual performance of the teacher and student models under homogeneous architecture settings.}
\label{tab:cifar_homo}
\small
\setlength{\tabcolsep}{4.5pt} 
\renewcommand{\arraystretch}{1.05} 

\resizebox{0.75\textwidth}{!}{
\begin{tabular}{l l c c c c c c}
\toprule
\multirow{4}{*}{Method} & \multirow{2}{*}{Teacher} & ResNet56 & ResNet110 & ResNet32x4 & WRN-40-2 & WRN-40-2 & VGG13 \\
 & & 72.40 & 74.31 & 79.42 & 75.59 & 75.59 & 74.64 \\
 & \multirow{2}{*}{Student} & ResNet20 & ResNet32 & ResNet8x4 & WRN-16-2 & WRN-40-1 & VGG8 \\
 & & 65.79 & 67.92 & 72.48 & 71.12 & 69.54 & 68.19 \\
\midrule
\multirow{5}{*}{Feature} & FitNet [29] & 69.64 & 72.35 & 75.26 & 74.46 & 73.28 & 70.09 \\
 & CRD & 67.28 & 71.97 & 74.16 & 74.22 & 72.44 & 72.74 \\
 & ReviewKD & 69.54 & 71.12 & 75.71 & 73.92 & 74.50 & 72.29 \\
 & CD & 70.78 & 73.66 & 74.46 & 74.45 & 71.34 & 75.61 \\
 & CoCaRS & \underline{72.40} & 74.45 & 76.24 & 75.20 & 75.22 & 74.56 \\
\midrule
\multirow{8}{*}{Logit} & KD  & 71.74 & 74.01 & 74.75 & 76.04 & 74.52 & 74.08 \\
 & DKD  & 71.17 & 74.12 & \underline{76.51} & 76.41 & \underline{75.33} & 74.41 \\
 & ABKD & 71.42 & 73.87 & 75.12 & 76.35 & 75.13 & 75.33 \\
 & WASB & 72.01 & 74.02 & 76.34 & \underline{76.45} & 74.88 & 74.45 \\
 & SKD & 71.54 & \underline{74.77} & 74.86 & 76.24 & 75.24 & \underline{75.75} \\
 & ACD (Ours) & \textbf{72.97} & \textbf{75.14} & \textbf{78.33} & \textbf{77.94} & \textbf{76.04} & \textbf{80.52} \\
\bottomrule
\end{tabular}
}
\end{table*}

\subsection{Angular Coordinate Interface (ACI)}

\label{subsec:aci}
To provide a geometry-compatible space for knowledge transfer, we reconstruct the student's classifier. Note that the teacher remains a standard linear classifier, ensuring our method requires no modification to the pre-trained teacher.

\subsubsection{Cosine Classifier}
We replace the student's linear classification head with a cosine classifier. The angular logit is computed as:
\begin{equation}
    \tilde{z}_k = s \cdot \cos\theta_k = s \cdot \frac{\mathbf{w}_k^\top \mathbf{h}}{\|\mathbf{w}_k\|_2 \|\mathbf{h}\|_2},
\end{equation}
where \(s\) is a global scale parameter controlling the sharpness of the distribution. By normalizing both the feature and class weights, ACI explicitly strips away the sample-dependent magnitudes (\(\|\mathbf{h}\|\)) and class-specific weight norms (\(\|\mathbf{w}_k\|\)). Setting the bias \(b_k=0\) further removes class-specific preferences. Consequently, the student classifier is restricted to a pure angular space, isolating the transferable directional semantic relations.

\subsubsection{Equiangular Tight Frame (ETF) Initialization}
While ACI establishes a pure angular space, the initialization of the class directions significantly impacts the early training dynamics. To provide a symmetric and uniformly distributed geometric prior, we initialize the angular classifier weights using an Equiangular Tight Frame (ETF).

Specifically, given \(C\) total classes and feature dimension \(d\) (\(d \ge C\)), the ETF prototype matrix \(\mathbf{W}_{\text{ETF}} = [\mathbf{w}_1, \dots, \mathbf{w}_C] \in \mathbb{R}^{d \times C}\) is constructed as:
\begin{equation}
    \mathbf{W}_{\text{ETF}} = \sqrt{\frac{C}{C-1}} \mathbf{U} \left( \mathbf{I}_C - \frac{1}{C} \mathbf{1}\mathbf{1}^\top \right),
\end{equation}
where \(\mathbf{U} \in \mathbb{R}^{d \times C}\) is a semi-orthogonal matrix satisfying \(\mathbf{U}^\top \mathbf{U} = \mathbf{I}_C\), \(\mathbf{I}_C\) is the identity matrix, and \(\mathbf{1}\) is the all-ones vector.

This initialization ensures that all class weight vectors have an equal \(l_2\) norm and identical pairwise angles. Specifically, the inner product between any two class directions satisfies:
\begin{equation}
    \mathbf{w}_i^\top \mathbf{w}_j = 
    \begin{cases} 
    1, & i = j, \\ 
    -\frac{1}{C-1}, & i \neq j. 
    \end{cases}
\end{equation}
By initializing the ACI classifier with \(\mathbf{W}_{\text{ETF}}\), the student model starts from a maximally separated angular configuration (a simplex). This geometric prior perfectly aligns with our SICD objective, which aims to transfer conditional non-target relations, providing a robust and well-conditioned space for subsequent distillation.

\begin{table*}[t]
\centering
\caption{Comparison of Top-1 accuracy on CIFAR-100. We report the individual performance of the teacher and student models under heterogeneous architecture settings.}
\label{tab:cifar_hetero}
\small
\setlength{\tabcolsep}{4pt} 
\renewcommand{\arraystretch}{1.05}

\resizebox{0.70\textwidth}{!}{
\begin{tabular}{l l c c c c c}
\toprule
\multirow{4}{*}{Method} & \multirow{2}{*}{Teacher} & ResNet32x4 & WRN-40-2 & VGG13 & ResNet50 & ResNet32x4 \\
& & 79.42 & 75.59 & 74.64 & 79.33 & 79.42 \\
& \multirow{2}{*}{Student} & ShuffleNetV1 & ShuffleNetV1 & MobileNetV1 & MobileNetV1 & ShuffleNetV2 \\
& & 69.43 & 69.43 & 60.09 & 60.09 & 73.5 \\
\midrule
\multirow{5}{*}{Feature} & FitNet [29] & 74.09 & 73.49 & 64.98 & 64.06 & 75.42 \\
& CRD & 75.50 & 76.15 & 66.55 & 66.28 & 74.66 \\
& ReviewKD & 77.39 & 77.68 & 65.26 & 62.52 & 77.99 \\
& CD & \underline{77.95} & \underline{77.85} & 64.98 & 67.97 & 78.13 \\
& CoCaRS & 77.85 & 76.01 & 68.12 & \underline{68.90} &\underline{78.53} \\
\midrule
\multirow{7}{*}{Logit} & KD [16] & 76.64 & 77.23 & 67.57 & 68.44 & 77.78 \\
& DKD & 76.75 & 75.94 & 67.58 & 67.51 & 78.43 \\
& ABKD & 77.01 & 75.14 & 66.56 & 67.81 & 77.87 \\
& WASB & 76.89 & 77.10 & 65.13 & 65.35 & 77.56 \\
& SKD & 77.91 & 77.53 & \underline{68.60} & 68.48 & 77.02 \\
& ACD (Ours) & \textbf{81.46} & \textbf{81.62} & \textbf{73.30} & \textbf{73.30} & \textbf{83.12} \\
\bottomrule
\end{tabular}
}
\end{table*}

\subsection{Scale-Invariant Conditional Distillation(SICD)} 

\subsubsection{Dual-Standardization}
To eliminate residual scale and shift effects, we apply Z-score standardization to the logits of both the teacher and the student before softmax:
\begin{equation}
    \hat{z}_k = \frac{z_k - \mu(\mathbf{z})}{\sigma(\mathbf{z})},
\end{equation}
where \(\mu(\mathbf{z})\) and \(\sigma(\mathbf{z})\) are the mean and standard deviation across all classes for a single sample. This step ensures that the supervision focuses purely on the relative class rankings, ignoring the absolute output magnitudes.

\subsubsection{Non-Target Conditional Distillation}
To prevent the student from mimicking the teacher's absolute confidence in the ground-truth class \(y\), we decompose the KL divergence into a binary confidence term and a conditional non-target relation term. We discard the former and distill only the latter:
\begin{equation}
    \mathcal{L}_{\mathrm{SICD}}
    = \tau^2 (1-p_y)\,
      D_{\mathrm{KL}}\!\bigl(
      p^{-y} \,\|\, q^{-y}
      \bigr).
    \label{eq:sicd}
\end{equation}
where $p^{-y}$ and $q^{-y}$ denote the teacher and student
conditional distributions over non-target classes.
For $k\neq y$, their entries are $p_k/(1-p_y)$ and
$q_k/(1-q_y)$, respectively.
This objective transfers relative probabilities among
non-target classes while omitting explicit
target-versus-non-target probability mass matching,
leaving cross-entropy to provide ground-truth supervision.

\subsubsection{Gradient Norm Matching (Enhancement)}
To stabilize the optimization, we introduce a gradient norm matching technique. The final SICD loss is scaled by a detached ratio \(r\) between the gradient magnitude of the full KD loss and the SICD loss:
\begin{equation}
    r = \frac{\|\nabla_{\theta} \mathcal{L}_{\text{KD}}\|}{\|\nabla_{\theta} \mathcal{L}_{\text{SICD}}\|},
\label{eq:r}
\end{equation}
where \(\theta\) represents the student backbone parameters. This normalization ensures that the distillation gradients remain comparable in scale to standard KD, preventing the SICD loss from being overwhelmed by the \(\mathcal{L}_{\text{CE}}\) term during early training.

\subsection{Overall Objective}
\label{subsec:overall}
By integrating the Angular Coordinate Interface (ACI) and the Scale-Invariant Conditional Distillation (SICD), the final training objective of our ACD framework is formulated as:
\begin{equation}
    \mathcal{L}_{\text{ACD}} = \mathcal{L}_{\text{CE}} + \lambda \cdot r \cdot \mathcal{L}_{\text{SICD}},
\end{equation}
where \(\mathcal{L}_{\text{CE}}\) is the standard cross-entropy loss for the classification task, \(\lambda\) is a balancing hyper-parameter, and \(r\) is the detached gradient norm matching ratio defined in Eq.~\ref{eq:r}.

%% file: sec/5_exp.tex
\section{Experiments}
\label{sec:experiments}

\subsection{Experimental Setup}
To assess the effectiveness of our method, we consider two types of teacher-student architecture pairs: (1) homogeneous pairs using the same architectures (e.g., VGG16 → VGG8), and (2) heterogeneous pairs involving different architectures (e.g., ResNet50 → MobileNetV1).  We included a range of architectures covering different capacity scales, such as ResNet, WideResNet (WRN), ShuffleNetV1/V2, MobileNetV1, and VGG.
We compare ACD against logit-based methods (KD~\cite{hintonDistillingKnowledgeNeural2015}, DKD~\cite{chenDistillingKnowledgeKnowledge}, ABKD~\cite{wangABKDPursuingProper2025}, WASB~\cite{naWASDWassersteinbasedKnowledge2026}, SKD~\cite{jeongStreamlinedKnowledgeDistillation} ) and feature/relation-based methods (FitNet~\cite{romeroFitNetsHintsThin2015}, CRD~\cite{tianContrastiveRepresentationDistillation2022}, ReviewedKD~\cite{yuanRevisitingKnowledgeDistillation}, CD~\cite{kangWhatStudentsLearn2026}, CoCaRS~\cite{yuCoCaRSCorrelationCalibrationBased2026}). 

\paragraph{Training Details.}
Experiments run in FP32 on NVIDIA A100 80GB GPUs.
For CIFAR-100, students are trained using SGD for 240 epochs with a batch size of 64, momentum of 0.9, and weight decay of $5\times10^{-4}$.
For ImageNet-1K, students are trained for 100 epochs using SGD with a batch size of 256, an initial learning rate of 0.05, momentum of 0.9, and weight decay of $10^{-4}$. The learning rate is reduced by a factor of 0.1 at epochs 30, 60, and 90.

\begin{table*}[t]
  \centering
  \caption{Comparison of Top-1 accuracy on ImageNet with both homogeneous (e.g., R34 $\to$ R18) and heterogeneous (e.g., R50 $\to$ MV1) teacher-student architectures.}
  \label{tab:imagenet_skd}
  \setlength{\tabcolsep}{4pt}
  \renewcommand{\arraystretch}{1.15}
  \resizebox{\textwidth}{!}{%
  \begin{tabular}{llccccccccc}
    \toprule
    Teacher & Student
    & KD~\cite{hintonDistillingKnowledgeNeural2015} & DKD~\cite{chenDistillingKnowledgeKnowledge} & AT~\cite{zagoruykoPayingMoreAttention2017} & ReviewKD~\cite{yuanRevisitingKnowledgeDistillation} & RKD~\cite{parkRelationalKnowledgeDistillation2019} & SKD~\cite{jeongStreamlinedKnowledgeDistillation} & CD~\cite{kangWhatStudentsLearn2026} & ACD (Ours)\\
    \midrule
    R34 (73.30) & R18 (69.76)
    & $69.11\pm0.21$
    & $70.88\pm0.16$
    & $70.54\pm0.15$
    & $70.51\pm0.18$
    & $69.94\pm0.20$
    & $71.13\pm0.12$
    & $71.01\pm0.10$
    & $\mathbf{71.64\pm0.12}$\\
    R50 (76.14) & MV1 (66.51)
    & $67.81\pm0.24$
    & $70.58\pm0.18$
    & $70.90\pm0.16$
    & $67.77\pm0.23$
    & $71.22\pm0.13$
    & $71.53\pm0.11$
    & $71.23\pm0.15$
    & $\mathbf{71.84\pm0.14}$\\
    \bottomrule
  \end{tabular}%
  }
\end{table*}

\begin{table*}[t]

  \centering
  \caption{Object detection results on MS-COCO using
  Faster R-CNN-FPN.}
  \label{tab:coco_skd}
  \setlength{\tabcolsep}{4pt}
  \renewcommand{\arraystretch}{1.15}
  \resizebox{0.70\textwidth}{!}{%
  \begin{tabular}{clccccccccc}
    \toprule
    \multirow{5}{*}{Method}
    & 
    & AP & AP$_{50}$ & AP$_{75}$
    & AP & AP$_{50}$ & AP$_{75}$
    & AP & AP$_{50}$ & AP$_{75}$ \\
    \cmidrule(lr){3-5}
    \cmidrule(lr){6-8}
    \cmidrule(lr){9-11}
    & \multirow{2}{*}{Teacher}
    & \multicolumn{3}{c}{ResNet101}
    & \multicolumn{3}{c}{ResNet101}
    & \multicolumn{3}{c}{ResNet50} \\
    &
    & 42.04 & 62.48 & 45.88
    & 42.04 & 62.48 & 45.88
    & 40.22 & 61.02 & 43.81 \\
    & \multirow{2}{*}{Student}
    & \multicolumn{3}{c}{ResNet18}
    & \multicolumn{3}{c}{ResNet50}
    & \multicolumn{3}{c}{MobileNetV2} \\
    &
    & 33.26 & 53.61 & 35.26
    & 37.93 & 58.84 & 41.05
    & 29.47 & 48.87 & 30.90 \\
    \midrule
    \multirow{3}{*}{Feature}
    & FitNet~\cite{romeroFitNetsHintsThin2015}
    & 34.13 & 54.16 & 36.71
    & 38.76 & 59.62 & 41.80
    & 30.20 & 49.80 & 31.69 \\
    & ReviewKD~\cite{yuanRevisitingKnowledgeDistillation}
    & 36.75 & 56.72 & 34.00
    & 40.36 & 60.97 & 44.08
    & 33.71 & 53.15 & 36.13 \\
    & MDPD~\cite{zhangMemoryefficientTransferLearning2026}
    & 36.14 & 56.51 & 38.37
    & 40.44 & 61.27 & 44.04
    & 31.75 & 50.89 & 32.34 \\
    \midrule
    \multirow{4}{*}{Logit}
    & KD~\cite{hintonDistillingKnowledgeNeural2015}
    & 33.97 & 54.66 & 36.62
    & 38.35 & 59.41 & 41.71
    & 30.13 & 50.28 & 31.35 \\
    & DKD~\cite{chenDistillingKnowledgeKnowledge}
    & 35.05 & 56.60 & 37.54
    & 39.25 & 60.90 & 42.73
    & 32.34 & 53.77 & 34.01 \\
    & SKD~\cite{jeongStreamlinedKnowledgeDistillation}
    & 36.74 & 57.03 & \textbf{39.76}
    & \textbf{40.62} & 61.61 & \textbf{44.61}
    & \textbf{34.25} & 54.51 & \textbf{36.29} \\
    & ACD (Ours)
    & \textbf{36.91} & \textbf{57.36} & 39.51
    & 40.17 & \textbf{61.67} & 43.79
    & 33.98 & \textbf{54.65} & 36.22 \\
    \bottomrule
  \end{tabular}%
  }
\end{table*}

\begin{table}[t]
    \centering
    \caption{\textbf{Component ablation.}
    Accuracy (\%); best results are bold.
    Aff.: affine head; CI: cosine interface;
    ACI: CI with equiangular tight frame (ETF) initialization;
    DS: dual standardization with full KL;
    GM: gradient matching.
    V13/V8: VGG13/VGG8; MNV1: MobileNetV1.}
    \label{tab:component_ablation}

    \small
    \setlength{\tabcolsep}{2pt}
    \setlength{\aboverulesep}{1.5pt}
    \setlength{\belowrulesep}{1.5pt}
    \renewcommand{\arraystretch}{1.0}

    \begin{tabular*}{\linewidth}{
        @{\extracolsep{\fill}}ccc@{}
    }
        \toprule
        Configuration
        & V13 $\to$ V8
        & V13 $\to$ MNV1 \\
        \midrule
        KD (Aff.)
        & $74.39 \pm 0.43$
        & $67.78 \pm 0.40$ \\

        CI
        & $75.10 \pm 0.19$
        & $66.60 \pm 0.16$ \\

        CI + ETF (ACI)
        & $75.80 \pm 0.15$
        & $68.10 \pm 0.14$ \\

        ACI + DS
        & $78.55 \pm 0.05$
        & $70.98 \pm 0.02$ \\

        ACI + SICD + GM
        & $\boldsymbol{80.52 \pm 0.02}$
        & $\boldsymbol{73.30 \pm 0.01}$ \\
        \bottomrule
    \end{tabular*}
\end{table}

\begin{table}[t]
    \centering
    \caption{\textbf{Compatibility with feature-based
    distillation on CIFAR-100.}
    Top-1 accuracy (\%).
    $\Delta$: gain over KD in percentage points.
    Best results within each row are bold.
    T/S: teacher/student;
    R32$\times$4: ResNet32$\times$4;
    SV1: ShuffleNetV1; V13: VGG13;
    MNV1: MobileNetV1.}
    \label{tab:feature_plugin}

    \small
    \setlength{\tabcolsep}{3pt}
    \setlength{\aboverulesep}{1.5pt}
    \setlength{\belowrulesep}{1.5pt}
    \renewcommand{\arraystretch}{1.05}

    \resizebox{\linewidth}{!}{%
        \begin{tabular}{@{}cccccc@{}}
            \toprule
            Method & T $\to$ S & KD & ACI & ACI + SICD & $\Delta$ \\
            \midrule
            AT~\cite{zagoruykoPayingMoreAttention2017}
            & R32$\times$4 $\to$ SV1
            & 77.30 & 77.60 & \textbf{78.97} & +1.67 \\

            ReviewKD~\cite{yuanRevisitingKnowledgeDistillation}
            & V13 $\to$ MNV1
            & 68.48 & 67.48 & \textbf{70.43} & +1.95 \\
            \bottomrule
        \end{tabular}%
    }
\end{table}

\subsection{Main Results}
We evaluate ACD on CIFAR-100, ImageNet, and MS-COCO
under homogeneous and heterogeneous teacher--student
architecture settings.
On CIFAR-100, Table~\ref{tab:cifar_homo} shows that ACD
outperforms prior logit-based and feature-based methods
across all six homogeneous pairs, improving over ordinary
KD by 2.51 percentage points on average.
Under heterogeneous settings
(Table~\ref{tab:cifar_hetero}), ACD achieves the highest
accuracy on all five pairs, exceeding the strongest
competing methods by 4.19 points on average.
On ImageNet (Table~\ref{tab:imagenet_skd}), ACD reaches
71.64\% and 71.84\% for
ResNet34$\rightarrow$ResNet18 and
ResNet50$\rightarrow$MobileNetV1, respectively,
surpassing the strongest baselines by 0.51 and 0.31 points.
For object detection on MS-COCO
(Table~\ref{tab:coco_skd}), ACD remains competitive:
it achieves the highest AP for
ResNet101$\rightarrow$ResNet18 and the highest AP$_{50}$
across all three pairs, while SKD retains stronger AP
on the other two pairs.
Together, these results demonstrate consistent
classification improvements and competitive transfer
to object detection

\begin{figure}[t]
  \centering
  \includegraphics[width=\linewidth]{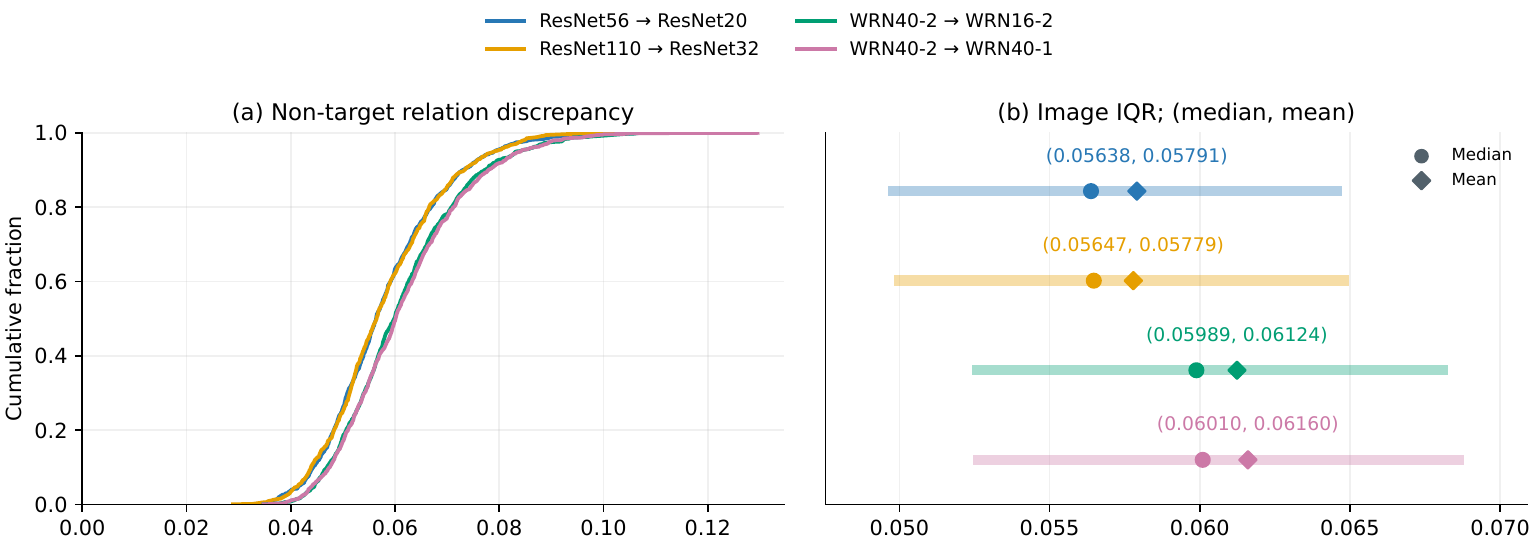}
  \caption{\textbf{Non-target relation discrepancies.}
  (a) Empirical cumulative distributions of conditional TV.
  (b) Medians (circles), means (diamonds), and image
  interquartile ranges (segments).}
  \label{fig:relation_preservation}
\end{figure}

\subsection{Ablation Studies}

Table~\ref{tab:component_ablation} evaluates the contributions
of ACI, ETF initialization, logit standardization, and SICD
on VGG13$\rightarrow$VGG8 and
VGG13$\rightarrow$MobileNetV1.
Replacing the affine head with ACI improves VGG8 by
0.71 percentage points but reduces MobileNetV1 accuracy
by 1.18 points, indicating that an angular interface alone
does not consistently improve distillation.
ETF initialization subsequently increases accuracy by
0.70 and 1.50 points, while standardized full KL provides
further gains of 2.75 and 2.88 points.
The complete ACI+SICD configuration reaches 80.52\%
and 73.30\%, improving over ordinary KD by 6.13 and
5.52 points, respectively.
Relative to the preceding standardized-full configuration,
the gains are 1.97 and 2.32 points.
This final comparison evaluates conditional supervision
together with gradient-strength matching, rather than
isolating either component.

\begin{figure}[t]
    \centering
    \includegraphics[width=\linewidth]{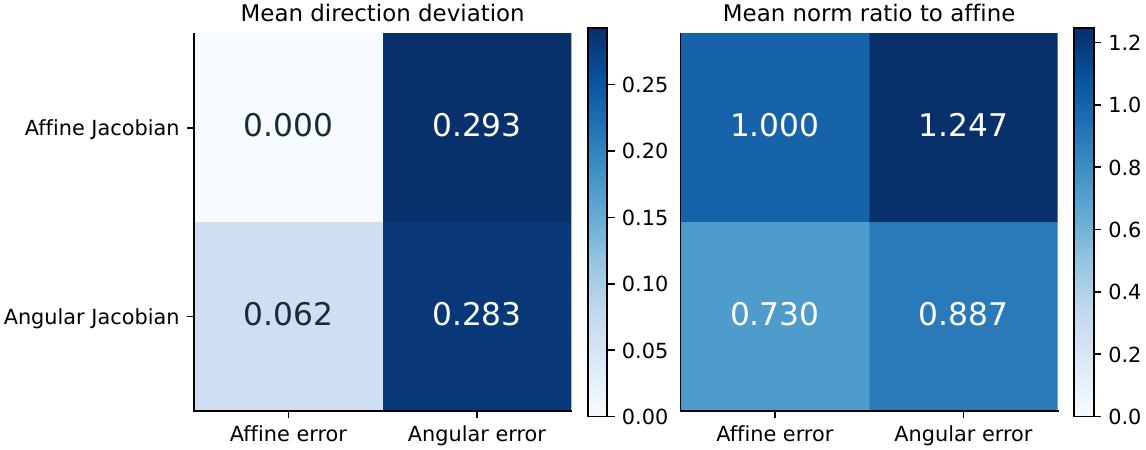}
    \caption{\textbf{Sources of KD feature-gradient changes.}}
    \label{fig:gradient_sources}
\end{figure}

\subsection{Further Analysis}

\paragraph{Gradient pathways of the angular interface.}
For the temperature-scaled KL objective, the gradient
with respect to student features $\mathbf{h}$ is
\begin{equation}
    g_h = J_h^\top T(q_T-p_T),
    \qquad
    J_h = \frac{\partial z}{\partial \mathbf{h}},
    \label{eq:kd_feature_gradient}
\end{equation}
where $p_T$ and $q_T$ denote the teacher and student
distributions at temperature $T$.
The angular interface can therefore affect both the
probability error and the classifier Jacobian.
At a fixed ImageNet CE-trained checkpoint, we cross
the affine and angular Jacobians with the two
corresponding probability-error vectors.
Relative to the original affine gradient, replacing only
the probability error, only the Jacobian, or both yields
mean direction deviations of 0.293, 0.062, and 0.283,
respectively (Fig.~\ref{fig:gradient_sources}).
The corresponding mean gradient-norm ratios are
1.247, 0.730, and 0.887.
These fixed-state diagnostics expose two interacting
paths through which the angular interface changes
the local distillation signal; the crossed combinations
are diagnostic constructions rather than trained variants.

\paragraph{Non-target relation agreement.}
We evaluate four completed ACI+SICD models on the same
1,024 CIFAR-100 test images.
After independently standardizing teacher and student
logits, we measure total variation between their
non-target conditional distributions at $T=4$.
Figure~\ref{fig:relation_preservation} presents the
image-level discrepancy distributions and their summary
statistics.
Mean conditional TV ranges from 0.0578 to 0.0616,
with similar distributions within the two ResNet pairs
and within the two WRN pairs.
These measurements quantify relation agreement at the
trained endpoints, rather than improvements over
matched baseline models.

\paragraph{Compatibility with feature-based distillation.}
We examine compatibility with two selected feature-distillation
pipelines by retaining their feature losses and adapters
while comparing affine + KL, cosine + KL, and cosine + SICD.
As shown in Table~\ref{tab:feature_plugin}, cosine + SICD
achieves 78.97\% with AT and 70.43\% with ReviewKD-style,
exceeding their affine + KL baselines by 1.67 and 1.95
percentage points, respectively.
Under the same cosine interface, replacing KL with SICD
yields gains of 1.37 and 2.95 percentage points.
The cosine interface alone reduces accuracy in the
ReviewKD-style pipeline, so its improvement cannot be
attributed to the classifier change alone.
These single-seed case studies support SICD as an
output-loss replacement within cosine-head
feature-distillation pipelines, while leaving the
feature-transfer components unchanged; they do not
establish universal gains across methods or architecture pairs.

\begin{figure}[t]
    \centering
    \includegraphics[width=0.8\linewidth]{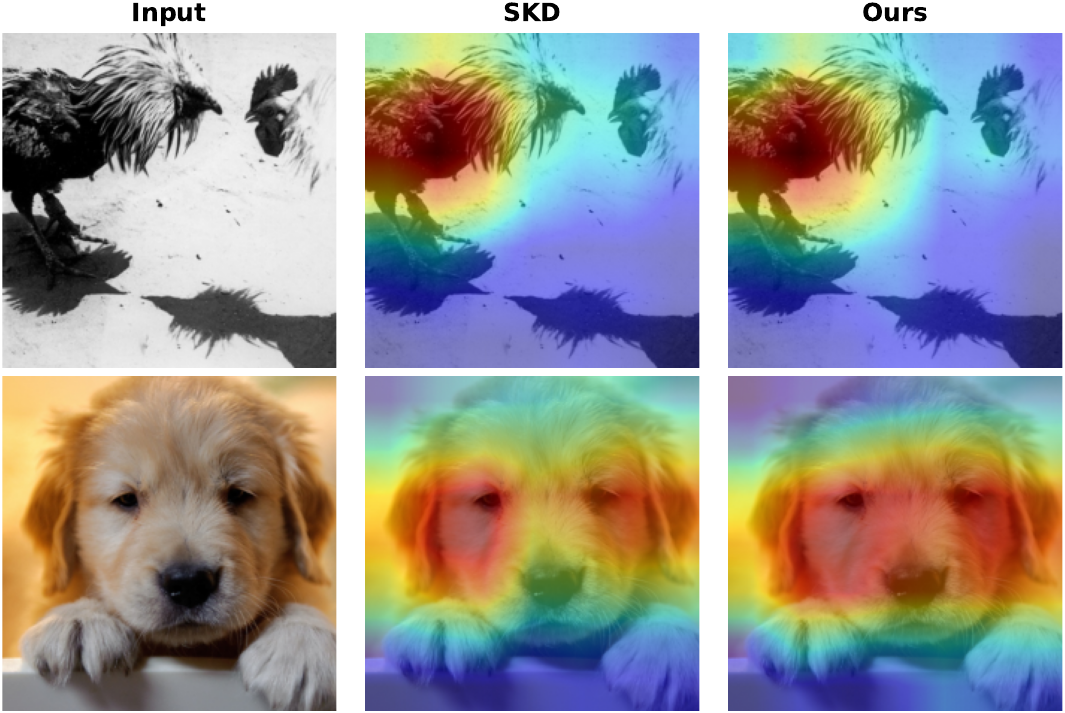}
    \caption{\textbf{Grad-CAM++ visualizations on ImageNet.}}
    \label{fig:gradcampp}
\end{figure}

\paragraph{Visualizing spatial attention.}
We complement the output and gradient diagnostics
with Grad-CAM++ visualizations on ImageNet.
Figure~\ref{fig:gradcampp} compares CE-trained and
KD-trained MobileNetV2 students using identical inputs,
ground-truth target classes, and the last convolutional
block before global pooling.
Both models highlight the foreground objects,
while exhibiting different activation patterns,
particularly around the dog's face.
These examples illustrate spatial differences between
the two training objectives; they do not yet evaluate
ACD or establish improved localization.

%% file: sec/6_conclusion.tex
\section{Conclusion}
\label{sec:conclusion}

We presented Angular Conditional Distillation (ACD), which
revisits knowledge distillation through the geometry of the
classifier interface.
ACI establishes a normalized,
bias-free cosine readout for the student, while SICD transfers
weighted non-target conditional relations through standardized
logits, omitting explicit target-mass matching.
Together, these components distinguish how class relations
are represented from which aspects of the teacher distribution
are distilled.
Experiments demonstrate the effectiveness of this design
on evaluated teacher--student pairs, while support its compatibility with existing
feature-transfer objectives.
Overall, our findings highlight classifier parameterization
and supervision selection as complementary considerations
in knowledge distillation.
Future work will investigate broader settings and more
efficient implementations of gradient matching.

\paragraph{LLM Usage Statement.}
Large language models were used to assist with language
editing, code development, and figure preparation.
All methodological decisions, experimental results,
and scientific claims were reviewed by the authors,
who take full responsibility for the final manuscript.